\documentclass[a4paper,conference]{IEEEtran}
\usepackage[latin9]{inputenc}
\usepackage{array}
\usepackage{float}
\usepackage{booktabs}
\usepackage{mathtools}
\usepackage{url}
\usepackage{multirow}
\usepackage{amsmath}
\usepackage{amssymb}
\usepackage{cancel}
\usepackage{mathdots}
\usepackage{stmaryrd}
\usepackage{stackrel}
\usepackage{undertilde}
\usepackage{graphicx}
\PassOptionsToPackage{version=3}{mhchem}
\usepackage{mhchem}
\usepackage{esint}
\usepackage[unicode=true,
 bookmarks=false,
 breaklinks=false,pdfborder={0 0 0},pdfborderstyle={},backref=false,colorlinks=false]
 {hyperref}

\makeatletter

\pdfpageheight\paperheight
\pdfpagewidth\paperwidth

\providecommand{\tabularnewline}{\\}
\floatstyle{ruled}
\newfloat{algorithm}{tbp}{loa}
\providecommand{\algorithmname}{Algorithm}
\floatname{algorithm}{\protect\algorithmname}

\ifCLASSINFOpdf
\else
\fi
\usepackage{algpseudocode}

\usepackage{setspace}

\makeatother

\begin{document}
\title{Explain2Attack: Text Adversarial Attacks via Cross-Domain Interpretability}
\author{\IEEEauthorblockN{Mahmoud Hossam, Trung Le, He Zhao, and Dinh Phung}
\IEEEauthorblockA{Clayton School of Information Technology, \\
Monash University, Victoria, Australia\\
 Emails: \{mhossam, trunglm, ethan.zhao, dinh.phung\}@monash.edu}  }
\maketitle
\begin{abstract}
Training robust deep learning models for down-stream tasks is a critical
challenge. Research has shown that down-stream models can be easily
fooled with adversarial inputs that look like the training data, but
slightly perturbed, in a way imperceptible to humans. Understanding
the behavior of natural language models under these attacks is crucial
to better defend these models against such attacks. In the black-box
attack setting, where no access to model parameters is available,
the attacker can only query the output information from the targeted
model to craft a successful attack. Current black-box state-of-the-art
models are costly in both computational complexity and number of queries
needed to craft successful adversarial examples. For real world scenarios,
the number of queries is critical, where less queries are desired
to avoid suspicion towards an attacking agent. In this paper, we propose
Explain2Attack, a black-box adversarial attack on text classification
task. Instead of searching for important words to be perturbed by
querying the target model, Explain2Attack employs an interpretable
substitute model from a similar domain to learn word importance scores.
We show that our framework either achieves or out-performs attack
rates of the state-of-the-art models, yet with lower queries cost
and higher efficiency.
\end{abstract}

\IEEEpeerreviewmaketitle{}

\section{Introduction}

Robustness in machine learning models is a critical challenge. Research
has shown that common downstream deep learning models can be easily
fooled with malicious input that looks like the training data, but
slightly perturbed, in a way imperceptible to humans. These perturbed
inputs are called adversarial examples, which can be used to attack
trained models, causing significant deterioration to down-stream task
performance. There has been a lot of work on generating adversarial
examples for different types of data, including images and text.
The better we understand how a model is vulnerable to different attacks,
the better we can increase its robustness. For instance, augmenting
crafted adversarial examples in the training data can improve robustness
of models \cite{goodfellow2014explaining}.

In general, attacks using adversarial examples can be crafted in either
white-box or black-box settings. In white-box attacks, the attacker
has access to the target model parameters, and the gradient of these
parameters is used to craft adversarial examples \cite{DBLP:journals/corr/abs-1812-08951,DBLP:journals/corr/abs-1902-07285,yang2019greedy,ijcai2018-601}.
On the other hand, black-box attacks do not have access to the model
parameters \cite{kuleshov2018adversarial,gao2018black,DBLP:journals/corr/abs-1907-11932},
but only to its outputs. In this paper, we are interested in black-box
attacks, since in practice, this is a more probable scenario for real-world
applications. 

Specifically, we consider in this paper black-box attacks on natural
language classification task. Typical classification models such
as deep neural networks (DNN) with softmax decision layer output a
probability distribution of their input belonging to each target class.
Usually, the final label of the model is decided to be the one with
the maximum probability. Hence, a classification model could be fooled
if the confidence of the output probability is affected by a malicious
input, switching the maximum probability to another incorrect target
class.

The key strategy used to craft adversarial text in existing methods
is to try to replace few words in an input sentence with synonyms
such that its meaning remains the same. The classification model is
then queried with these perturbed sentences to find out which ones
successfully changes the output label. Existing state-of-the-art models
have different ways to search for most important words to replace,
but the common intuition is to compute the importance score for each
word as a function of the probability output of target model (see
Section \ref{sec:W2_Literature-Review} for further details).

Since existing approaches rely on word by word querying of the target
model, they are costly in both computational complexity and number
of queries. For real world scenarios, the number of queries is critical,
where less queries are desired to avoid suspicion towards an attacking
agent.

In this paper, we propose \textit{Explain2Attack}\footnote{Code is available at: \url{https://github.com/mahossam/Explain2Attack}},
a black-box adversarial attack on text classification, that employs
cross-domain interpretability to learn word importance for crafting
adversarial examples. The key idea is to replace the need to querying
the target model by learning a similar substitute model with similar
domain data, that can then be used to generate word importance scores
for the targeted model. The advantages of our model are: \textit{(i)
less costly in computational complexity and number of queries, (ii)
achieves or out-performs state-of-the-art methods in attack rates,
yet with less number of queries, and (iii) has better scalability
with longer input lengths compared to current methods.}

\section{Background}

Here we formally define the adversarial attack problem, and the details
regarding adversarial crafting process. We discuss related work in
details compared to our work in Section \ref{sec:W2_Literature-Review}.

\subsection{Problem Definition \label{subsec:Problem-Definition}}

 Let $D$ be a dataset of $N$ sentences and corresponding labels
$D={\left\{ \mathbb{X},\mathbb{Y}\right\} }$, where $\mathbb{X}=\left\{ X_{1},X_{2},...,X_{N}\right\} $
is a corpus of $N$ sentences, and $\mathbb{Y}=\left\{ {Y_{1},Y_{2},...,Y_{N}}\right\} $
is the collection of the class labels of $M$ possible text classes.
A pre-trained target model $F:\mathbb{X}\rightarrow\mathbb{Y}$ is
the classifier model we want to attack. $F$ maps the input space
$\mathbb{X}$ to the label space $\mathbb{Y}$. Starting from an original
sentence $X\in\mathbb{X}$, a valid adversarial example $X_{adv}$
could be crafted such that: 
\begin{equation}
F(X_{adv})\neq F(X),\text{{and}}\,Sim(X_{adv},X)\ge\epsilon\label{eq:SucessAdv-1}
\end{equation}
where $Sim:\mathbb{X}\times\mathbb{X}\rightarrow(0,1)$ is a similarity
function and $\epsilon$ is the minimum desired similarity between
the original and adversarial examples. In the case of natural language,
this is usually a combination of semantic and syntactic similarity
\cite{DBLP:journals/corr/abs-1907-11932,vijayaraghavan2019generating}.

\subsection{Crafting Adversarial Examples}

To craft an adversarial example for a given sentence $X\in\mathbb{X}$
, the common strategy to follow is : i) selecting the most important
words/tokens to replace from the input sentence, then ii) searching
for synonyms to replace the most important words such that the changed
sentence changes the classification label of the target model. iii)
Finally, in order for the final adversarial example to be plausible
and imperceptible to humans, the semantic similarity between the original
candidate sentence and the final one need to be close to each other
or restricted using some sentence similarity function $Sim(\cdot,\cdot)$.

\subsection{Word Importance Ranking \label{subsec:Word-Importance-Ranking}}

Since the search space for all possible word placements to attack
sentence $X$ is large, most black-box attacks use a word importance
ranking criteria, that helps prioritize which words in $X$ to replace
first.

Let $I_{w_{i}}$ be a score to measure the influence of a word $w_{i}\in X$
towards the model output probability $F_{Y}(X)$ of the predicted
label $Y$. Different black-box methods differ on how to compute
$I_{w_{i}}$ for each word in a sentence, as discussed later in Section
\ref{sec:W2_Literature-Review}. However, they share the same need
for the probability output of the classifier for class label $Y$,
where $I_{w_{i}}$ is computed as a function of these probability
outputs: 
\begin{equation}
I_{w_{i}}=\text{{ScoreFunction}}(F_{Y}(X))\label{eq:fn_general_word_impotance_score}
\end{equation}

\subsection{Word Replacement}

After word ranking is done, word replacement step begins. In this
step, the algorithm takes candidate words by order of importance,
and replaces the current word by chosen synonyms till the target model
label is changed. The candidate sentence $X_{adv}$ with changed words
is considered a valid adversarial example if it passes the condition
in Eq.\ref{eq:SucessAdv-1}. After all word replacements are tried,
if $X_{adv}$ still could not change the label, then we assume that
no adversarial example can be crafted from sentence $X$.

\section{\label{sec:W2_Framework}Proposed Framework}

We propose a more efficient word ranking and selection model that
alleviates the need for output probabilities for word ranking, hence,
is more efficient in number of needed queries of target model. Our
approach is to build an interpretable substitute model that can closely
resemble the target model behaviour on the attacked data domain. Then,
using the interpretability capability of the substitute model, we
can produce importance scores that can benefit our attack task.

\begin{figure*}
\begin{centering}
\includegraphics[viewport=20bp 0bp 800bp 580bp,clip,scale=0.45]{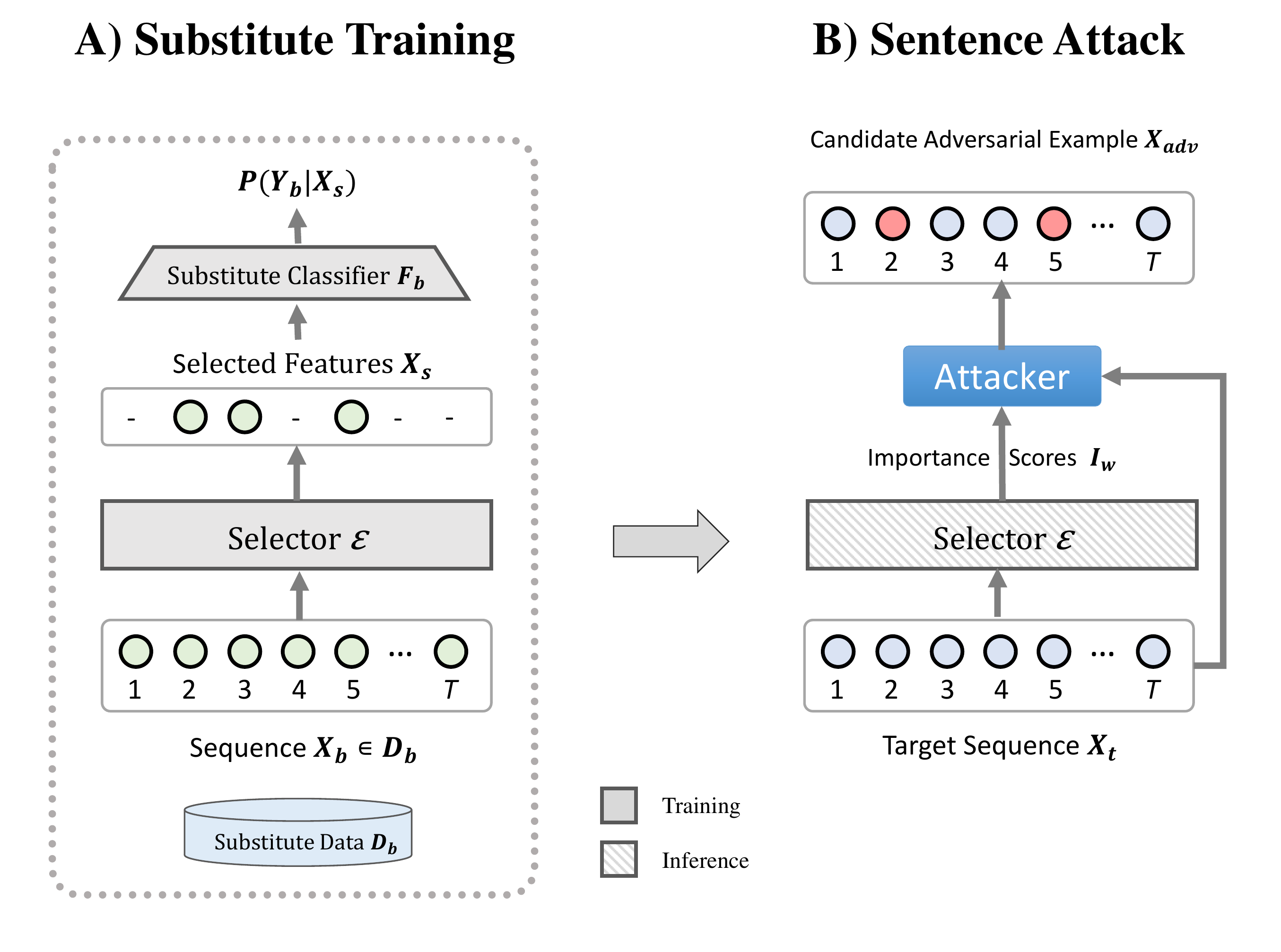}
\par\end{centering}
\caption{\label{fig:Overview-of-Explain2Attack}Overview of Explain2Attack}
\end{figure*}

In order to adapt our setup to work in black-box setting, where we
do not have access to attacked training data, we rely on the domain
adaption capacity of deep models. Potentially, there is a similarity
of language sentences representing a certain linguistic concept. For
example, the words and semantic structures of the reviews for a restaurant
and a movie can be usually similar, except that the subjects/objects
of the reviews are different. Therefore, if a model is trained to
capture such high-level features, the knowledge of the model can be
transferred between datasets. This means that, if we do not have access
to target model training dataset, we still can train a close enough
\textit{substitute} model on \textit{similar }dataset. In practice,
deep learning models exhibit domain adaptation capacity, where a
model trained on certain data, can still behave well on other similar
data that has similar high level features. Thus, attacks produced
for the proposed substitute model \cite{papernot2017practical} can
also be used to efficiently attack the original target model as long
as they both are trained from similar domains. In figure \ref{fig:Overview-of-Explain2Attack}
we show the overview of our method.

In details, consider a target model $F$ trained on some target training
dataset $D_{t}^{\textrm{train}}=\left\{ \mathbb{X}_{t}^{\textrm{train}},\mathbb{Y}_{t}^{\textrm{train}}\right\} $
and testing set $D_{t}^{\textrm{test}}=\left\{ \mathbb{X}_{t}^{\textrm{test}},\mathbb{Y}_{t}^{\textrm{test}}\right\} $.
Instead of querying target model $F(\cdot)$ with context around each
word in sentences from $\mathbb{X}_{t}^{\textrm{test}}$, we consider
learning a substitute interpretable model that can provide importance
scores for given words. For this purpose, we leverage a framework
extended from \cite{pmlr-v80-chen18j} to train the substitute model.

\subsection{Interpretable Substitute Model}

 In black-box attacks setup, we do not have access to target training
data $D_{t}^{\textrm{train}}$. Therefore, in order to train a substitute
model that can generate importance scores during attack, we need to
look for another dataset that is close enough to the target model
dataset. We call such a dataset the substitute dataset, $D_{b}=\left\{ \mathbb{X}_{b},\mathbb{Y}_{b}\right\} $.
We then use $D_{b}$ to train the substitute model we call $\textrm{SUB}$.

Our goal from training $\textrm{SUB}$ is to learn a network (called
the \textit{selector}) that can select the most important features
from the input $X\in\mathbb{X}_{b}$ to let another network, the \textit{substitute
classifier}, correctly predict the corresponding label $Y\in\mathbb{Y}_{b}$.
In details, inspired by \cite{pmlr-v80-chen18j}, for a substitute
pair $(X,Y)\sim D_{b}$, our goal is to learn a selector network $\mathcal{E}(X)$
that selects the most important subset of $k$ features from $X$
that is sufficient for substitute classifier $F_{b}(.)$ to correctly
predict $Y$. After substitute training is finished, $F_{b}(.)$ can
be discarded, since we are only interested in the selector $\mathcal{E}(X)$.

Formally, for a given positive integer $k$, let $\rho_{k}$ be the
set of all possible subsets of size $k$:

\[
\rho_{k}=\left\{ S\subset2^{d}\mid|S|=k\right\} 
\]
We denote the selected $k$ features indices as $S$, and the corresponding
selected features sub-vector from $X$ as $X_{s}$:

\[
S\mathbb{\sim P}(S\mid x)=\mathcal{E}(X)
\]
where $S\in\rho_{k}$ and $X_{S}\in\mathbb{R}^{k}$. The choice of
the number of explaining features $k$ can be tuned as a hyper-parameter.
The learning objective is to find $\mathcal{E}$ that maximizes the
mutual information $I\left(X_{S};Y\right)$:
\begin{equation}
\max_{\mathcal{E}}I\left(X_{S};Y\right)\quad\text{ subject to }\quad S\sim\mathcal{E}(X)\label{eq:L2X_approx}
\end{equation}

Since a direct solution to the Problem \ref{eq:L2X_approx} is not
tractable, an approximate solution can be found using a lower bound
on the mutual information. We refer the reader to \cite{pmlr-v80-chen18j}
for more details on the final lower bound objective.

In the attack step, in order to generate word importance scores from
the trained $\textrm{SUB}$, we take the logit output of the last
layer of $\mathcal{E}(X)$ before the $k$ selection process $S\sim\mathcal{E}(X)$.
This way we obtain importance score vector $I\in\mathbb{R}^{d}$ that
has the same dimensions of the input.

The selector $\mathcal{E}$ takes the form of a multilayer deep convolutional
neural network. In order for $\mathcal{E}$ to select $k$ features
from $X$, a Gumbel-Softmax layer \cite{DBLP:journals/corr/JangGP16}
is used to generate $k$ indices.

\subsection{Adversarial Examples via Cross-Domain Interpretability}

At inference time, when crafting the adversarial attacks, we do inference
on $\textrm{SUB}$ using the target testing set $D_{t}^{\textrm{test}}$.
This is the standard setup for black-box attack, where $D_{t}^{\textrm{test}}$
is used as starting point and slightly perturbed to craft valid adversarial
examples. As described earlier, because of the domain adaption capacity
of deep learning models, the closer the substitute domain is to the
target, the better the generated scores will be for the original target
model.

\subsection{Implementation}

Our attack method is inspired by framework proposed in \cite{DBLP:journals/corr/abs-1907-11932}.
However, we modified the behaviour so that we alleviate the need for
querying $F(\cdot)$ for word ranking, since our proposed method relies
on interpretability architecture for this purpose. We name our framework
\textit{Explain2Attack}. In Algorithms \ref{alg:Train-L2X-substitute}
and \ref{alg:Explain2Attack} we describe in details how our algorithm
works.

\begin{algorithm}[h]
\caption{\label{alg:Train-L2X-substitute}Train Substitute Model $\textrm{SUB}$}



\begin{algorithmic}
\item[] 
\Statex \textbf{Input:} Substitute training corpus $\mathbb{X}_{b}=\left\{ X_{1},X_{2},...,X_{N}\right\} $ of $N$ sentences, and corresponding class labels $\mathbb{Y}_{b}=\left\{ {Y_{1},Y_{2},...,Y_{N}}\right\} $
\Statex \textbf{Output:} Trained substitute model $\textrm{SUB}$ with selector network $\mathcal{E}$
\item[] 
\For{each mini batch $B_{s}\in\{\mathbb{X}_{b},\mathbb{Y}_{b}\}$}
\State	Update parameters of $\textrm{SUB}$ model to find $\mathcal{E}$ that {\par}maximizes an approximate to objective in Eq. (\ref{eq:L2X_approx})
\EndFor

\end{algorithmic}
\textbf{\small{}}{\small\par}
\end{algorithm}

{\footnotesize{}}
\begin{algorithm}[h]
{\footnotesize{}\caption{Explain2Attack{\small{}}}
}{\footnotesize\par}

{\footnotesize{}\label{alg:Explain2Attack}}{\footnotesize\par}


\begin{spacing}{0.95}
\begin{algorithmic}[1]
\Statex 
	\Statex \textbf{Input:} Target test sentence example $X_{t}^{\textrm{test}}\in \mathbb{X}_{t}^{\textrm{test}}$, where $X_{t}^{\textrm{test}}=\left\{ w_{1},w_{2},\ldots,w_{n}\right\} $, the corresponding ground truth label $Y_{t}^{\text{test}}$, target model $F$, sentence similarity function $Sim(\cdot,\cdot)$, sentence similarity threshold $\epsilon$, word embeddings $Emb$ over the vocabulary $\mathbf{Vocab}$
\Statex \textbf{Output:} Perturbed Adversarial example $X_{adv}$

	\item[]
		\State Initialization: $X_{adv}\leftarrow X_{t}^{\textrm{test}}$
		\For{each word index $\boldsymbol{i}$ in $X_{t}^{\textrm{test}}$}
				\State Retrieve word importance score: \par $I_{w_{i}}=$ {\footnotesize{}$\mathbf{Get\_Word\_Score}$}{\footnotesize} $(X_{t}^{\textrm{test}}, \boldsymbol{i})$ 
		
		\EndFor
		\item[]
		\State $W\leftarrow$ Sorted list of words in $w_{i}\in X$ in descending order of importance score $I_{w_{i}}$ 
		\item[]
		\For{each word $w_{j}\in W$}
			\State $\mathbf{Candidates}\leftarrow\{\,\}$ 
			\For{each nearest synonym word $c_{m} \in \mathbf{Vocab}$ (by \par embedding cosine similarity)}
				\State $X_{m}^{\prime}\leftarrow$ Replace $w_{j}$ with $c_{m}$ in $X_{adv}$ 
				\State Add $X_{m}^{\prime}$ to $\mathbf{Candidates}$
			\EndFor
		
		\item[]
		\If{$\exists \, X^{\prime}\in \mathbf{Candidates}$ s.t. $F(X^{\prime})\neq Y_{t}^{test}$ and \par $Sim\left(X_{t}^{\textrm{test}},X^{\prime}\right)\ge\epsilon$}		
			\State $X_{adv}\leftarrow $ $\underset{X^{\prime}\in \mathbf{Candidates}}{\textrm{argmax}}$ $Sim\left(X_{t}^{\textrm{test}},X^{\prime}\right)$		
			\Statex	
			\State \textbf{return} $X_{adv}$
			\Statex	
		\ElsIf{$F_{Y}\left(X_{adv}\right)>\underset{X^{\prime} \in \mathbf{Candidates}}{\min}F_{Y}(X^{\prime})$}
				\item[]
				\State $X_{adv}\leftarrow\underset{X^{\prime}\in \mathbf{Candidates}}{\textrm{argmin}}F_{Y}(X^{\prime})$
		\EndIf
		\item[]
		\EndFor
		\State \textbf{return} None
\item[]
\par
\Function{$\textrm{Get\_Word\_Score}$}{$X_{t},i$}
	\State $\textrm{word\_score}=\ensuremath{\mathcal{E_{\textrm{logits}}}}(X_{t})_{i}$ 
	\State \textbf{return} $\textrm{word\_score}$
\EndFunction
\end{algorithmic}
\end{spacing}

\end{algorithm}
{\footnotesize\par}

We follow Algorithm \ref{alg:Train-L2X-substitute} to train the substitute
model $\textrm{SUB}$ using $D_{b}$. After that, selector $\mathcal{E}$
from $\textrm{SUB}$ can be used in algorithm \ref{alg:Explain2Attack}
for crafting adversarial examples.

The procedure to craft adversarial examples is described in Algorithm
\ref{alg:Explain2Attack}. The algorithm starts from an input sentence
$X_{t}^{\textrm{test}}$ and terminates either after successfully
finding a perturbed adversarial example $X_{adv}$ that changes the
label $Y_{t}^{\textrm{test}}$, or if no such perturbation is found.
In details, the algorithm proceeds by inferring from $\textrm{\ensuremath{\mathcal{E}}}$
word scores for every word in $X_{t}^{\textrm{test}}$. These scores
are sorted, called $W$, and then the algorithm tries to replace word
by word to find valid $X_{adv}$. For each word $w_{j}$ in $W$ a
set of close candidate synonyms is selected based on embedding cosine
similarity as in \cite{DBLP:journals/corr/abs-1907-11932}, and a
set of sentences $\textrm{\textbf{Candidates}}$ is created by replacing
$w_{j}$ with each synonym.

At the current word $w_{j}$ , the algorithm first checks if perturbing
$w_{j}$ in $X_{adv}$ using the set of synonyms in $\textrm{\textbf{Candidates}}$
can change $Y_{t}^{\textrm{test}}$. If one or more such synonym are
found, the one which achieves the highest similarity $Sim\left(X_{t}^{\textrm{test}},X^{\prime}\right)$
to original sentence is picked, and the algorithm terminates.

Otherwise, if no synonym is found that can change $Y_{t}^{\textrm{test}}$,
the algorithm needs to choose the synonym perturbation that yields
the weakest (minimum) target model probability $F_{Y}(X^{\prime})$
before moving on to the next word $w_{j+1}$.

\section{\label{sec:W2_Literature-Review}Related Work}

There has been recent work on adversarial text attacks \cite{DBLP:journals/corr/abs-1812-08951,DBLP:journals/corr/abs-1902-07285,kuleshov2018adversarial,yang2019greedy,ijcai2018-601,gao2018black,ren-etal-2019-generating,DBLP:journals/corr/abs-1907-11932}
. The main challenges for natural language adversarial attacks is
discrete nature of inputs, where defining meaningful perturbations
is not straight forward, and the search space and complexity for black-box
attack methods.

Specifically for black-box text attacks, several methods \cite{DBLP:journals/corr/abs-1907-11932,ren-etal-2019-generating,gao2018black}
have been developed that share similar general framework, where the
attack starts by selecting the most important words/tokens to replace
from a candidate sentence, followed by searching for some word replacement
that can flip the classification label of the target model. However,
some methods followed the heuristic optimization approach, for example,
\cite{alzantot-etal-2018-generating} used genetic algorithm to find
the best sentence perturbation that fools the classifier.  

Most of formerly mentioned black-box methods use the word selection/replacement
strategy. For instance, PWWS \cite{ren-etal-2019-generating} proposes
computing a word saliency score using output probabilities of the
target model, while \cite{gao2018black} computes sequential importance
score based on forward and backward RNN probabilities at the current
word position in the sentence. TextFooler \cite{DBLP:journals/corr/abs-1907-11932}
is a recent strong baseline for text attacks, where the method uses
a modified procedure for word ranking that increases the ranking in
label disagreement case. BERT-Attack and BAE-Attack \cite{li2020bertattack,garg2020bae}
improve on TextFooler synonym replacement by using a pretrained language
model to generate suitable substitute words based on the surrounding
context. This achieved higher attack rates and the number of queries
are reduced. The improvement in BERT-Attack can be easily incorporated
in our framework.

Our approach differs from previous work in solving the word ranking
problem. Unlike other methods, instead of depending on the target
model for word importance ranking, we \textit{learn} word importance
scores. The main differences of our approach compared exiting ones
are: i) word ranking has no dependence on the target model output,
thus more efficient in number of queries, ii) unlike exiting methods,
our approach is scalable with increased sentence lengths, since computing
the scores is not dependent on word by word query of target model.
This makes our approach more efficient, scalable, and less computationally
expensive compared to existing methods. Moreover, our general approach
can benefit from further query reduction in the synonym replacement
phase by incorporating the pretrained language model technique in
BERT-Attack and BAE-Attack.

\section{\label{sec:W2_Experiments}Experiments}

We report here the results of our method on text classification tasks.
We apply our framework to several sentiment classification datasets
with WordCNN, WordLSTM and BERT \cite{devlin2018bert} target models.
However, our model can be applied to other classification models or
datasets with proper choice of substitute datasets . We compare our
results to TextFooler \cite{DBLP:journals/corr/abs-1907-11932}, a
strong state-of-the-art baseline for black-box text attack on the
chosen target models. Below we describe the datasets, metrics and
discuss the results. In Table \ref{tab:Explain2Attack_datasets} we
report the datasets we used in our experiment with their statistics.

\subsubsection*{Datasets}
\begin{itemize}
\item \textbf{IMDB} and \textbf{MR}: Movie reviews for sentiment classification
\cite{maas2011learning,pang2005seeing}. The reviews have binary labels,
either positive or negative.
\item \textbf{Amazon MR}: Amazon polarity (binary) user reviews on movies,
extracted from the larger Amazon reviews polarity dataset \footnote{\url{https://www.kaggle.com/bittlingmayer/amazonreviews}}.
\item \textbf{Yelp Polarity Reviews}: Sentiment classification on positive
and negative businesses reviews \cite{zhang2015character}. We mainly
use this dataset as a substitute dataset when attacking other models.
\end{itemize}
In all of the datasets except Amazon MR, we follow the data preprocessing
and partitioning in \cite{DBLP:journals/corr/abs-1907-11932}.

\begin{table}[h]
\caption{\label{tab:Explain2Attack_datasets}Statistics of Used Datasets}

\centering{}{\small{}}%
\begin{tabular}{c|ccc}
\hline 
\textbf{Dataset} & \textbf{Train} & \textbf{Test} & \textbf{Avg. Length}\tabularnewline
\hline 
IMDB & 25K & 25K & 215\tabularnewline
MR & 9K & 1K & 20\tabularnewline
Amazon MR & 25K & 25K & 100\tabularnewline
Yelp & 560K & 38K & 152\tabularnewline
\end{tabular}{\small{}}{\small\par}
\end{table}

\begin{table*}
\caption{\label{tab:test_Accuracy_Queries}After-Attack Accuracies, Queries
and Query Efficiency}

\resizebox{1\textwidth}{!}{
\begin{tabular}{cccc|ccc|ccc}
\hline 
\multicolumn{2}{c}{{\small{}Classifier}} & \multicolumn{2}{c}{{\small{}BERT}} & \multicolumn{3}{c}{{\small{}WordCNN}} & \multicolumn{3}{c}{{\small{}WordLSTM}}\tabularnewline
\hline 
\multicolumn{2}{c}{{\small{}Target Model}} & {\small{}IMDB} & {\small{}MR} & {\small{}IMDB} & {\small{}MR} & {\small{}Amazon MR} & {\small{}IMDB} & {\small{}MR} & {\small{}Amazon MR}\tabularnewline
\hline 
 & {\small{}$\textrm{Clean\_Acc}$.} & {\footnotesize{}92.18} & {\footnotesize{}89.97} & {\footnotesize{}87.32} & {\footnotesize{}79.85} & {\footnotesize{}90.14} & {\footnotesize{}88.78} & {\footnotesize{}81.82} & {\footnotesize{}91.30}\tabularnewline
\hline 
\multirow{3}{*}{{\small{}Adv\_Acc. $\downarrow$}} & {\small{}TextFooler \cite{DBLP:journals/corr/abs-1907-11932}} & {\footnotesize{}11.88} & {\footnotesize{}13.59} & \textbf{\footnotesize{}0.60} & {\footnotesize{}1.50} & \textbf{\footnotesize{}3.92} & \textbf{\footnotesize{}0.04} & \textbf{\footnotesize{}2.06} & \textbf{\footnotesize{}2.15}\tabularnewline
\cline{2-10} \cline{3-10} \cline{4-10} \cline{5-10} \cline{6-10} \cline{7-10} \cline{8-10} \cline{9-10} \cline{10-10} 
 & \textit{\footnotesize{}(Substitute Data)} & \textit{\footnotesize{}(Yelp)} & \textit{\footnotesize{}(Amazon MR)} & \multicolumn{1}{c|}{\textit{\footnotesize{}(Yelp)}} & \multicolumn{2}{c|}{\textit{\footnotesize{}(IMDB)}} & \multicolumn{2}{c|}{\textit{\footnotesize{}(Amazon MR)}} & \textit{\footnotesize{}(IMDB)}\tabularnewline
 & {\small{}Explain2Attack (ours)} & \textbf{\footnotesize{}11.32} & \textbf{\footnotesize{}13.34} & {\footnotesize{}0.61} & \textbf{\footnotesize{}1.31} & {\footnotesize{}3.97} & {\footnotesize{}0.06} & {\footnotesize{}2.27} & {\footnotesize{}2.38}\tabularnewline
\hline 
\multirow{2}{*}{{\small{}$\textrm{Avg\_Queries}$ $\downarrow$}} & {\small{}TextFooler} & {\footnotesize{}980.5} & \textbf{\footnotesize{}181.6} & {\footnotesize{}444} & {\footnotesize{}112.8} & {\footnotesize{}378.7} & {\footnotesize{}500.2} & {\footnotesize{}117.5} & {\footnotesize{}392.7}\tabularnewline
 & {\small{}Explain2Attack} & \textbf{\footnotesize{}873.5} & {\footnotesize{}184.07} & \textbf{\footnotesize{}404.5} & \textbf{\footnotesize{}108.7} & \textbf{\footnotesize{}349.4} & \textbf{\footnotesize{}440.5} & \textbf{\footnotesize{}114.2} & \textbf{\footnotesize{}369.3}\tabularnewline
\hline 
\multirow{2}{*}{{\small{}}%
\begin{tabular}{c}
{\small{}Query Efficiency}\tabularnewline
{\small{}($\textrm{QE}$)}\tabularnewline
\end{tabular}{\small{} $\uparrow$}} & {\small{}TextFooler} & {\footnotesize{}0.082} & \textbf{\footnotesize{}0.421} & {\footnotesize{}0.195} & {\footnotesize{}0.695} & {\footnotesize{}0.228} & {\footnotesize{}0.177} & {\footnotesize{}0.679} & {\footnotesize{}0.227}\tabularnewline
 & {\small{}Explain2Attack} & \textbf{\footnotesize{}0.093} & {\footnotesize{}0.416} & \textbf{\footnotesize{}0.214} & \textbf{\footnotesize{}0.723} & \textbf{\footnotesize{}0.247} & \textbf{\footnotesize{}0.201} & \textbf{\footnotesize{}0.697} & \textbf{\footnotesize{}0.241}\tabularnewline
\end{tabular}}
\end{table*}

\subsubsection*{Metrics}

We evaluate our method using the following metrics: 
\begin{itemize}
\item \textbf{After-attack accuracy} \textbf{(}$\textrm{Adv\_Acc}$\textbf{)}:
We report for each model the original clean accuracy $\textrm{Clean\_Acc}$
of the target model on the test set. Then we report the accuracy of
the target model against crafted adversarial examples $\textrm{Adv\_Acc}$.
The lower is this better, where the larger the gap between these two
accuracies means more successful the attack method. Through-out discussion,
when we refer to ``\textbf{\textit{attack rate}}'', we mean the
gap $(\textrm{Clean\_Acc}-\textrm{Adv\_Acc})$.
\item \textbf{Average number of queries (}$\textrm{Avg\_Queries}$\textbf{):}
We report the average number of queries needed to find successful
adversarial example per input sentence. The lower is better, where
lower number of queries is one of the main desired goals of our method.
This is an absolute measure of the number of queries regardless of
the achieved after-attack accuracy.
\item \textbf{Query Efficiency ($QE$): }Since more successful attacks need
more queries in black-box setting, we cannot rely only on $\textrm{Avg\_Queries}$
for evaluating the performance of our method. We need to measure the
true benefit in reduction of number of queries, and make sure that
it is not reduced because of sacrificing attack rate. This is the
motivation behind Query Efficiency ($QE$), the ratio of successful
attacks per query $QE=\frac{\textrm{Clean\_Acc}-\textrm{Adv\_Acc}}{\textrm{Avg\_Queries}}$.
The higher is better. This means that $\mathit{QE}$ measures the
percentage of successful attacks per single query, hence the true
query efficiency related to the attack rate.
\item \textbf{Perturbation Query Cost (}$PQC$\textbf{)}: The number of
queries needed per perturbed word $PQC=\frac{\textrm{Avg\_Queries}}{\textrm{Pertureb\_Words}}$.
The lower is better. $PQC$ measures the cost in terms of queries
needed for a useful word perturbation that contributes towards label
change.
\end{itemize}

\subsection*{Discussion }

The main focus in evaluating the performance of our method is the
number of queries, both as absolute measure and its efficiency relative
to the achieved attack rate. These two aspects of evaluation are critical
to measure the true gains we get from our approach.

In Table \ref{tab:test_Accuracy_Queries}, we compare after-attack
accuracies and corresponding average number of queries to the baseline.
For each target model and dataset, we report the best substitute dataset
that yielded the best result.  Across all but one case, we find that
our method either achieves or out-performs attack rates of the baseline,
yet with lower number of queries. The gains in terms of number of
queries differs according to different datasets. We discuss this in
details later in this section.

It is important to study the cases where our method achieves lower
number of queries but does not achieve higher attack rate than the
baseline. In the black-box setting, higher attack rate is related
to number of queries, since there is a cost of certain number of queries
for every word perturbation. This means that lower number of queries
is only beneficial if it improves or preserves the attack rate. Therefore,
in order to measure the true efficiency of our method in terms of
queries per successful attack, we report in Table \ref{tab:test_Accuracy_Queries}
Query Efficiency ratio $\mathit{QE}$. We find that when our method
has lower number of queries, it has better $\mathit{QE}$ ratio, even
if the attack rate is not better than the baseline. This implies that
the lower number of queries achieved comes from true efficiency of
our method, not because of corresponding drop in attack rate.

\begin{table*}[!t]
\caption{\label{tab:MR_Queries_Length}Effect of Sentence Length on Number
of Queries}

\centering{}{\small{}}%
\begin{tabular}{c|c|ccc|cc|ccc}
\hline 
\multicolumn{2}{c|}{{\small{}Target Dataset}} & \multicolumn{3}{c|}{{\small{}IMDB}} & \multicolumn{2}{c|}{{\small{}Amazon MR}} & \multicolumn{3}{c}{{\small{}MR}}\tabularnewline
\hline 
\multicolumn{2}{c|}{{\small{}Classifier}} & {\small{}BERT} & {\small{}CNN} & {\small{}LSTM} & {\small{}CNN} & {\small{}LSTM} & {\small{}BERT} & {\small{}CNN} & {\small{}LSTM}\tabularnewline
\hline 
\multicolumn{2}{c|}{{\small{}Average Sentence Length}} & \multicolumn{3}{c|}{{\small{}215}} & \multicolumn{2}{c|}{{\small{}100}} & \multicolumn{3}{c}{{\small{}20}}\tabularnewline
\hline 
\multirow{2}{*}{{\small{}$\textrm{Avg\_Queries}$ $\downarrow$}} & {\small{}TextFooler} & {\small{}980.5} & {\small{}444} & {\small{}500.2} & {\small{}378.7} & {\small{}392.7} & {\small{}112.8} & {\small{}117.5} & \textbf{\small{}181.6}\tabularnewline
 & {\small{}Explain2Attack} & \textbf{\small{}873.5} & \textbf{\small{}404.5} & \textbf{\small{}440.5} & \textbf{\small{}349.4} & \textbf{\small{}369.3} & \textbf{\small{}108.7} & \textbf{\small{}114.2} & {\small{}184.07}\tabularnewline
\hline 
\multicolumn{2}{c|}{{\small{}Difference}} & {\small{}106.5} & {\small{}39.5} & {\small{}59.7} & {\small{}29.3} & {\small{}23.4} & {\small{}4.1} & {\small{}3.3} & {\small{}-3.0}\tabularnewline
\end{tabular}{\small\par}
\end{table*}

\begin{table*}[!t]
\caption{\label{tab:queries_per_word}Perturbation Query Cost\textbf{ ($\textrm{PQC}$)}}

\resizebox{1\textwidth}{!}{%
\centering{}{\small{}}%
\begin{tabular}{cccc|ccc|ccc}
\hline 
\multicolumn{2}{c}{{\small{}Classifier}} & \multicolumn{2}{c}{{\small{}BERT}} & \multicolumn{3}{c}{{\small{}WordCNN}} & \multicolumn{3}{c}{{\small{}WordLSTM}}\tabularnewline
\hline 
\multicolumn{2}{c}{{\small{}Target Model}} & {\small{}IMDB} & {\small{}MR} & {\small{}IMDB} & {\small{}MR} & {\small{}Amazon MR} & {\small{}IMDB} & {\small{}MR} & {\small{}Amazon MR}\tabularnewline
\hline 
\multirow{3}{*}{{\small{}Max Perturbed Words}} & {\small{}TextFooler} & {\footnotesize{}222} & {\footnotesize{}20} & {\footnotesize{}56} & {\footnotesize{}14} & \textbf{\footnotesize{}35} & {\footnotesize{}89} & \textbf{\footnotesize{}13} & {\footnotesize{}75}\tabularnewline
\cline{2-10} \cline{3-10} \cline{4-10} \cline{5-10} \cline{6-10} \cline{7-10} \cline{8-10} \cline{9-10} \cline{10-10} 
 & \textit{\footnotesize{}(Substitute Data)} & \textit{\footnotesize{}(Yelp)} & \textit{\footnotesize{}(Amazon MR)} & \multicolumn{1}{c|}{\textit{\footnotesize{}(Yelp)}} & \multicolumn{2}{c|}{\textit{\footnotesize{}(IMDB)}} & \multicolumn{2}{c|}{\textit{\footnotesize{}(Amazon MR)}} & \textit{\footnotesize{}(IMDB)}\tabularnewline
 & {\small{}Explain2Attack} & \textbf{\footnotesize{}127} & {\footnotesize{}20} & \textbf{\footnotesize{}54.3} & \textbf{\footnotesize{}11} & {\footnotesize{}41.7} & \textbf{\footnotesize{}84.7} & {\footnotesize{}13.8} & \textbf{\footnotesize{}68}\tabularnewline
\hline 
\multirow{2}{*}{{\small{}Avg. Perturbed Words}} & {\small{}TextFooler} & {\footnotesize{}18.7} & {\footnotesize{}4.2} & {\footnotesize{}5.8} & {\footnotesize{}2.3} & {\footnotesize{}5.0} & {\footnotesize{}7.6} & {\footnotesize{}2.5} & {\footnotesize{}7.0}\tabularnewline
 & {\small{}Explain2Attack} & {\footnotesize{}22} & {\footnotesize{}4.8} & {\footnotesize{}9.1} & {\footnotesize{}2.8} & {\footnotesize{}6.5} & {\footnotesize{}11.3} & {\footnotesize{}2.9} & {\footnotesize{}8.7}\tabularnewline
\hline 
\multirow{2}{*}{{\small{}}%
\begin{tabular}{c}
{\small{}Perturbation Query Cost}\tabularnewline
{\small{}($\textrm{PQC}$)}\tabularnewline
\end{tabular}{\small{}$\downarrow$}} & {\small{}TextFooler} & {\footnotesize{}52.5} & {\footnotesize{}43.5} & {\footnotesize{}76.6} & {\footnotesize{}49.0} & {\footnotesize{}76.3} & {\footnotesize{}65.8} & {\footnotesize{}47.0} & {\footnotesize{}56.4}\tabularnewline
 & {\small{}Explain2Attack} & \textbf{\footnotesize{}39.7} & \textbf{\footnotesize{}38.6} & \textbf{\footnotesize{}44.5} & \textbf{\footnotesize{}38.8} & \textbf{\footnotesize{}53.8} & \textbf{\footnotesize{}39.2} & \textbf{\footnotesize{}39.2} & \textbf{\footnotesize{}42.2}\tabularnewline
\end{tabular}}
\end{table*}

\begin{table*}
\caption{\label{tab:example_sentences}Adversarial Examples Sentences. Perturbed
Words are Highlighted}

\centering{}\resizebox{1\textwidth}{!}{
\begin{tabular}{cll}
\toprule 
 & Label & Sentence\tabularnewline
\midrule
Original & (0=Negative) & the film lapses too often into sugary \textbf{sentiment} and withholds
delivery on the pell mell pyrotechnics\tabularnewline
 &  & its punchy style promises\tabularnewline
Adversarial & (1=Positive) & the film lapses too often into sugary \textbf{emotions} and withholds
delivery on the pell mell pyrotechnics\tabularnewline
 &  & its punchy style promises\tabularnewline
\midrule
Original & (1=Positive) & the movie sticks much closer to hornby 's drop dead confessional tone
than the film version of high \textbf{fidelity} did\tabularnewline
Adversarial & (0=Negative) & the movie sticks much closer to hornby 's drop dead confessional tone
than the film version of high \textbf{faithful} did\tabularnewline
\midrule
Original & (0=Negative) & i'm not exactly sure what this \textbf{movie thinks} it is about\tabularnewline
Adversarial & (1=Positive) & i'm not exactly sure what this \textbf{cinematography concepts} it
is about\tabularnewline
\midrule
Original & (1=Positive) & the performances of the four \textbf{main} actresses bring their characters
to life a \textbf{little} melodramatic\tabularnewline
 &  & , but with enough \textbf{hope} to keep you engaged\tabularnewline
Adversarial & (0=Negative) & the performances of the four \textbf{underlying} actresses bring their
characters to life a \textbf{littlest} melodramatic\tabularnewline
 &  & , but with enough \textbf{wanting} to keep you engaged\tabularnewline
\end{tabular}}
\end{table*}

\paragraph*{Effect of Sentence Length}

In the case of MR dataset in Table \ref{tab:test_Accuracy_Queries},
we find that the reduction in number of queries is the lowest among
other datasets. By comparing the statistics of the datasets (Table
\ref{tab:Explain2Attack_datasets}) with the results in Table \ref{tab:test_Accuracy_Queries},
we find a correlation between the dataset average sentence length
and the corresponding reduction in attack queries. The longer the
sentences are, the more reduction our method achieves in number of
queries. We summarize this finding in Table \ref{tab:MR_Queries_Length}.
This effect is in agreement with our design goals, since unlike other
existing methods, our method does not perform word by word ranking
through target model querying. Hence, with longer input sentences,
the re duction in number of queries is improved by our method. This
is a key advantage of our method in terms of scalability to datasets
with longer sentences.

\paragraph*{Query Cost of Perturbed Words}

Another important measure is the number of queries needed for every
word perturbation; Perturbation Query Cost\textbf{ (}$PQC$\textbf{)}.
We are interested in this measure in order to understand the added
cost of queries if the model requires more words to be perturbed to
find a successful attack. In Table \ref{tab:queries_per_word} we
report $PQC$ against the baseline. In addition, we report both the
maximum and average number of perturbed words needed for successful
adversarial attack. Results show that our method out-performs the
baseline in $PQC$ measure, meaning that our method scales better
with number of perturbations needed for successful attacks.

\paragraph*{Qualitative Assessment}

We show in Table \ref{tab:example_sentences} examples of successful
adversarial examples from our method. We find that the language semantic
is preserved, and that the choice of perturbed words resemble important
keywords that contribute to the original label.

\paragraph*{Transferability Conditions}

In all of our experiments, we used the standard L2X \cite{pmlr-v80-chen18j}
architecture as the choice for the substitute architecture. However,
we found that changing the substitute architecture affects attack
rates. This finding is similar to \cite{demontis2019adversarial,madry2018towards}
where attack performance is found to be related to the model's architecture
and complexity. We further look to study in details why and when attacks
transfer well between substitute and target models. However, we leave
such comprehensive study for future work.

\section{Conclusion}

We propose a general framework that employs interpretability and domain
transfer for crafting black-box text adversarial attacks. The main
intuition is to learn word ranking that most probably impacts the
target model instead of searching for it expensively. To achieve this,
we train an interpretable substitute model on a substitute dataset,
with no need for the target dataset. Thus, the substitute model learns
importance scores with less number of queries and higher efficiency.
Results show that our method reduces queries cost for attacking text
classification models, while achieving or out-performing the state-of-the-art
attack rate. We show that our method is superior in both query cost
of perturbations and with longer input sentences, which allows our
method to scale better with longer sentences.

For future work, we plan to overcome the current limitations in our
current framework. Specifically, we plan to \textit{i)} train the
selector network to directly fool the target classifier by leveraging
its output information, in order to improve attack rate, \textit{ii)}
study and formalize the transferability conditions from substitute
to target domains, and provide guides for choice of suitable substitute
models, and \textit{iii)} further reduce dependence on target model
by learning replacement synonyms through a pretrained language model.

\bibliographystyle{IEEEtran}
\bibliography{references}

\begin{thebibliography}{10}
\providecommand{\url}[1]{#1}
\csname url@samestyle\endcsname
\providecommand{\newblock}{\relax}
\providecommand{\bibinfo}[2]{#2}
\providecommand{\BIBentrySTDinterwordspacing}{\spaceskip=0pt\relax}
\providecommand{\BIBentryALTinterwordstretchfactor}{4}
\providecommand{\BIBentryALTinterwordspacing}{\spaceskip=\fontdimen2\font plus
\BIBentryALTinterwordstretchfactor\fontdimen3\font minus
  \fontdimen4\font\relax}
\providecommand{\BIBforeignlanguage}[2]{{%
\expandafter\ifx\csname l@#1\endcsname\relax
\typeout{** WARNING: IEEEtran.bst: No hyphenation pattern has been}%
\typeout{** loaded for the language `#1'. Using the pattern for}%
\typeout{** the default language instead.}%
\else
\language=\csname l@#1\endcsname
\fi
#2}}
\providecommand{\BIBdecl}{\relax}
\BIBdecl

\bibitem{goodfellow2014explaining}
I.~J. Goodfellow, J.~Shlens, and C.~Szegedy, ``Explaining and harnessing
  adversarial examples,'' 2014.

\bibitem{DBLP:journals/corr/abs-1812-08951}
\BIBentryALTinterwordspacing
Y.~Belinkov and J.~R. Glass, ``Analysis methods in neural language processing:
  {A} survey,'' \emph{CoRR}, vol. abs/1812.08951, 2018. [Online]. Available:
  \url{http://arxiv.org/abs/1812.08951}
\BIBentrySTDinterwordspacing

\bibitem{DBLP:journals/corr/abs-1902-07285}
\BIBentryALTinterwordspacing
W.~Wang, B.~Tang, R.~Wang, L.~Wang, and A.~Ye, ``Towards a robust deep neural
  network in texts: A survey,'' \emph{CoRR}, vol. abs/1902.07285, 2019.
  [Online]. Available: \url{http://arxiv.org/abs/1902.07285}
\BIBentrySTDinterwordspacing

\bibitem{yang2019greedy}
P.~Yang, J.~Chen, C.-J. Hsieh, J.-L. Wang, and M.~I. Jordan, ``Greedy attack
  and gumbel attack: Generating adversarial examples for discrete data,'' 2019.

\bibitem{ijcai2018-601}
M.~Sato, J.~Suzuki, H.~Shindo, and Y.~Matsumoto, ``Interpretable adversarial
  perturbation in input embedding space for text,'' in \emph{Proceedings of the
  Twenty-Seventh International Joint Conference on Artificial Intelligence,
  {IJCAI-18}}.\hskip 1em plus 0.5em minus 0.4em\relax International Joint
  Conferences on Artificial Intelligence Organization, 7 2018, pp. 4323--4330.

\bibitem{kuleshov2018adversarial}
V.~Kuleshov, S.~Thakoor, T.~Lau, and S.~Ermon, ``Adversarial examples for
  natural language classification problems,'' 2018.

\bibitem{gao2018black}
J.~Gao, J.~Lanchantin, M.~L. Soffa, and Y.~Qi, ``Black-box generation of
  adversarial text sequences to evade deep learning classifiers,'' in
  \emph{2018 IEEE Security and Privacy Workshops (SPW)}.\hskip 1em plus 0.5em
  minus 0.4em\relax IEEE, 2018, pp. 50--56.

\bibitem{DBLP:journals/corr/abs-1907-11932}
D.~Jin, Z.~Jin, J.~T. Zhou, and P.~Szolovits, ``Is {BERT} really robust?
  natural language attack on text classification and entailment,'' \emph{CoRR},
  vol. abs/1907.11932, 2019.

\bibitem{vijayaraghavan2019generating}
P.~Vijayaraghavan and D.~Roy, ``Generating black-box adversarial examples for
  text classifiers using a deep reinforced model,'' 2019.

\bibitem{papernot2017practical}
N.~Papernot, P.~McDaniel, I.~Goodfellow, S.~Jha, Z.~B. Celik, and A.~Swami,
  ``Practical black-box attacks against machine learning,'' in
  \emph{Proceedings of the 2017 ACM on Asia conference on computer and
  communications security}, 2017, pp. 506--519.

\bibitem{pmlr-v80-chen18j}
J.~Chen, L.~Song, M.~Wainwright, and M.~Jordan, ``Learning to explain: An
  information-theoretic perspective on model interpretation,'' in
  \emph{Proceedings of the 35th International Conference on Machine Learning},
  ser. Proceedings of Machine Learning Research, vol.~80.\hskip 1em plus 0.5em
  minus 0.4em\relax PMLR, 10--15 Jul 2018, pp. 883--892.

\bibitem{DBLP:journals/corr/JangGP16}
\BIBentryALTinterwordspacing
E.~Jang, S.~Gu, and B.~Poole, ``Categorical reparameterization with
  {Gumbel-Softmax},'' \emph{ArXiv e-prints}, vol. abs/1611.01144, 2016.
  [Online]. Available: \url{http://arxiv.org/abs/1611.01144}
\BIBentrySTDinterwordspacing

\bibitem{ren-etal-2019-generating}
S.~Ren, Y.~Deng, K.~He, and W.~Che, ``Generating natural language adversarial
  examples through probability weighted word saliency,'' in \emph{Proceedings
  of the 57th Annual Meeting of the Association for Computational
  Linguistics}.\hskip 1em plus 0.5em minus 0.4em\relax Florence, Italy:
  Association for Computational Linguistics, Jul. 2019, pp. 1085--1097.

\bibitem{alzantot-etal-2018-generating}
\BIBentryALTinterwordspacing
M.~Alzantot, Y.~Sharma, A.~Elgohary, B.-J. Ho, M.~Srivastava, and K.-W. Chang,
  ``Generating natural language adversarial examples,'' in \emph{Proceedings of
  the 2018 Conference on Empirical Methods in Natural Language
  Processing}.\hskip 1em plus 0.5em minus 0.4em\relax Brussels, Belgium:
  Association for Computational Linguistics, Oct.-Nov. 2018, pp. 2890--2896.
  [Online]. Available: \url{https://www.aclweb.org/anthology/D18-1316}
\BIBentrySTDinterwordspacing

\bibitem{li2020bertattack}
L.~Li, R.~Ma, Q.~Guo, X.~Xue, and X.~Qiu, ``Bert-attack: Adversarial attack
  against bert using bert,'' 2020.

\bibitem{garg2020bae}
S.~Garg and G.~Ramakrishnan, ``Bae: Bert-based adversarial examples for text
  classification,'' 2020.

\bibitem{devlin2018bert}
J.~Devlin, M.-W. Chang, K.~Lee, and K.~Toutanova, ``Bert: Pre-training of deep
  bidirectional transformers for language understanding,'' \emph{arXiv preprint
  arXiv:1810.04805}, 2018.

\bibitem{maas2011learning}
A.~Maas, R.~E. Daly, P.~T. Pham, D.~Huang, A.~Y. Ng, and C.~Potts, ``Learning
  word vectors for sentiment analysis,'' in \emph{Proceedings of the 49th
  annual meeting of the association for computational linguistics: Human
  language technologies}, 2011, pp. 142--150.

\bibitem{pang2005seeing}
B.~Pang and L.~Lee, ``Seeing stars: Exploiting class relationships for
  sentiment categorization with respect to rating scales,'' \emph{arXiv
  preprint cs/0506075}, 2005.

\bibitem{zhang2015character}
X.~Zhang, J.~Zhao, and Y.~LeCun, ``Character-level convolutional networks for
  text classification,'' in \emph{Advances in neural information processing
  systems}, 2015, pp. 649--657.

\bibitem{demontis2019adversarial}
A.~Demontis, M.~Melis, M.~Pintor, J.~Matthew, B.~Biggio, O.~Alina, N.-R.
  Cristina, and F.~Roli, ``Why do adversarial attacks transfer? explaining
  transferability of evasion and poisoning attacks,'' in \emph{28th
  $\{$USENIX$\}$ Security Symposium ($\{$USENIX$\}$ Security 19)},
  vol.~28.\hskip 1em plus 0.5em minus 0.4em\relax $\{$USENIX$\}$ Association,
  2019.

\bibitem{madry2018towards}
A.~Madry, A.~Makelov, L.~Schmidt, D.~Tsipras, and A.~Vladu, ``Towards deep
  learning models resistant to adversarial attacks,'' in \emph{International
  Conference on Learning Representations}, 2018.

\end{thebibliography}

\end{document}